\documentclass[10pt, conference, compsocconf]{IEEEtran}
\newcommand{\remove}[1]{}
\usepackage{amsmath}
\usepackage{amssymb}
\usepackage{gensymb}
\usepackage{hyperref}

\newcommand{\eqn}[1]{Eqn.$\,$(\ref{eqn:#1})}

\newcommand{\mysection}[1]{Section \ref{sec:#1}}
\def\bea{\begin{eqnarray}}
\def\eea{\end{eqnarray}}
\newcommand{\fig}[1]{Fig. \ref{fig:#1}}

\def\be{\begin{equation}}
\def\ee{\end{equation}}
\def\bea{\begin{eqnarray}}
\def\eea{\end{eqnarray}}

\newcommand{\cD}{\mathcal{D}}
\newcommand{\cS}{\mathcal{S}}
\newcommand{\cX}{\mathcal{X}}
\newcommand{\cY}{\mathcal{Y}}

\newcommand{\ud}[1]{\, \mathrm{d}#1}

\newcommand{\EXP}{\mathbb{E}}

\usepackage{cite}
\usepackage{booktabs}

\usepackage{bibspacing}
\usepackage{lineno,xcolor}
\usepackage{multicol}

\usepackage{url}
\usepackage{wrapfig}
\usepackage[font=footnotesize]{subfig}
\usepackage{multirow}
\usepackage{array}
\newcolumntype{P}[1]{>{\centering\arraybackslash}p{#1}}
\usepackage[capitalize]{cleveref}

\usepackage{epsfig}
\usepackage{color}
\usepackage{amsmath}
\usepackage{amssymb}
\usepackage{gensymb}

\usepackage{subfig}

\usepackage{graphicx}
\usepackage{algorithm,algpseudocode}
\usepackage[]{caption}

\usepackage{draftwatermark}
\usepackage{comment}
\usepackage{cancel}
\usepackage[font=scriptsize,labelfont=bf]{caption}
\SetWatermarkAngle{55}
\SetWatermarkLightness{0.85}
\SetWatermarkFontSize{8cm}
\SetWatermarkText{}
\SetWatermarkScale{0.225}
\IEEEoverridecommandlockouts
\begin{document}
%-----------------------------------------------------------------------------

%\pagenumbering{arabic}
%-----------------------------------------------------------------------------
\title
{An Active Learning-Based Streaming  Pipeline for Reduced Data Training of Structure Finding Models in Neutron Diffractometry}

%-----------------------------------------------------------------------------

%-----------------------------------------------------------------------------
\author{
%\remove{
  \IEEEauthorblockN{Tianle Wang$^a$, Jorge Ramirez$^b$, Cristina Garcia-Cardona$^c$, Thomas Proffen$^d$, Shantenu Jha$^{e,f,g}$ and Sudip K. Seal$^b$}
   $^a$Computational Science Initiative, %\\  
  Brookhaven National Laboratory, USA\\
  $^b$Computer Science and Mathematics Division, %\\  
  Oak Ridge National Laboratory, USA\\
  $^c$Computer, Computational and Statistical Sciences Division, %\\  
  Los Alamos National Laboratory, USA\\
  $^d$Spallation Neutron Source, %\\  
  Oak Ridge National Laboratory, USA\\
  $^e$Rutgers, The State University of New Jersey, USA\\
  $^f$Princeton Plasma Physics Laboratory, USA\\
  $^g$Princeton University, USA
  %}
  \thanks{
  This manuscript has been authored by UT-Battelle, LLC under Contract No.
  DE-AC05-00OR22725 with the U.S. Department of Energy. The United States
  Government retains and the publisher, by accepting the article for publication,
  acknowledges that the United States Government retains a non-exclusive, paid-up,
  irrevocable, worldwide license to publish or reproduce the published form of
  this manuscript, or allow others to do so, for United States Government
  purposes. The Department of Energy will provide public access to these results
  of federally sponsored research in accordance with the DOE Public Access Plan
  (http://energy.gov/downloads/doe-public-access-plan)
  %
  %A portion of this research used resources at the Spallation Neutron Source, a DOE  Office of Science User Facility operated by the Oak Ridge National Laboratory.
  %
  This research used resources at the Argonne Leadership Computing Facility, the National Energy Research Scientific Computing Center and the Spallation Neutron Source, which are DOE  Office of Science User Facilities as well as in the Oak Ridge National Laboratory and the Brookhaven National Laboratory which are DOE Office of Science National Laboratories.
  This research was sponsored by the ExaLearn Co-Design Project, an Exascale Computing Project, DOE.
  }
}
%-----------------------------------------------------------------------------
\maketitle
\thispagestyle{plain}
\pagestyle{plain}

%-----------------------------------------------------------------------------
\begin{abstract}
Structure determination workloads in neutron diffractometry are computationally expensive and routinely require several hours to many days to determine the structure of a material from its neutron diffraction patterns. The potential for machine learning models  trained on simulated neutron scattering patterns to significantly speed up these tasks have been  reported recently. However, the amount of simulated data needed to train these models grows exponentially with the number of structural parameters to be predicted and poses a significant computational challenge. 
%To overcome this challenge, we introduce a novel batch-mode active learning policy that uses uncertainty sampling to simulate training data drawn from a probability distribution that prefers labelled examples about which the model is least certain. We discuss the design of a streaming training workflow that uses this active learning policy and present a performance study on two heterogeneous CPU+GPU platforms that confirms its efficacy in training the same models with $\sim 75\%$ less training data  in $\sim 20\%$ shorter training time and without any loss of accuracy. 
To overcome this challenge, we introduce a novel batch-mode active learning (AL) policy that uses uncertainty sampling to simulate training data drawn from a probability distribution that prefers labelled examples about which the model is least certain. We confirm its efficacy in training the same models with $\sim 75\%$ less training data while improving the accuracy. We then discuss the design of an efficient stream-based training workflow that uses this AL policy and present a performance study on two heterogeneous platforms to demonstrate that, compared with a conventional training workflow, the streaming workflow delivers $\sim 20\%$ shorter training time without any loss of accuracy.
\end{abstract}
%-----------------------------------------------------------------------------

%%%%%%%%%%%%%%%%%%%%%%%%%%%%%%%%%%%%
\section{Introduction}
\label{sec:intro}
The structure of crystalline solids is characterized by repeating arrangements of atoms, ions or molecules. Determining the structure of a crystalline material involves computing the structure of the smallest unit of these repeating patterns, defined by three {\em unit cell lengths} $\{a,b,c\}$ and three {\em unit cell angles} $\{\alpha,\beta,\gamma\}$. Neutron diffraction experiments or {\em neutron diffractometry} is a state-of-the-art method to study these structural parameters. In nature, the unit cell lengths and unit cell angles are constrained to satisfy unique relations between themselves which, in turn, define the distinct crystallographic symmetry classes to which the unit cells belong. 
The task of determining the structure of any crystalline material, therefore, reduces to that of determining the unit cell lengths/angles and the specific relation 
they satisfy. The task of identifying which symmetry class a material belongs to is a {\em classification} task while that of determining the cell lengths and angles is a {\em regression} task.

Conventional approaches for computing the unit cell parameters, $\{a,b,c,\alpha,\beta,\gamma\}$, use a loop refinement method.  In this approach, a forward physics model simulates the neutron scattering pattern, known as a {\em Bragg profile}, based on an initial guess for the cell parameters. The simulated pattern is then compared with the observed pattern using a pre-defined measure of similarity. If the similarity falls within a specified tolerance threshold, the guessed cell parameters used as inputs to the forward model are accepted as the structural parameters of the material under study. Otherwise, the process is repeated with a new set of input cell parameters until the threshold of similarity tolerance is met. This iterative loop-refinement process is computationally intensive, especially when a high-fidelity forward model is used, often requiring hours, days, and even weeks to complete.
To address this challenge, it was recently shown in \cite{Cardona-Big-Data-2020} that, once trained, ML models have the potential to predict the structural parameters directly from their Bragg profiles with high accuracy and in a fraction of the time needed by refinement methods. 

Training supervised ML models for crystalline structure identification is data intensive due to the high resolution required in the prediction of unit cell parameters, where differences in length of 0.01 angstroms or in angles of 0.15$\degree$ could distinguish between distinct crystallographic symmetries. Examples are typically generated by sampling from a structured grid of possible unit-cell parameters. The construction of such grids is also symmetry-dependent. A more constrained symmetry such as {\em cubic}, requires the prediction of only one lattice parameter, the length $a$ (all the lengths are equal $a=b=c$ and all the angles are equal and known $\alpha = \beta = \gamma = 90\degree$), and therefore, needs exhaustive stepping through a one-dimensional space, i.e. simulating for length values $a$ within a given range with small enough step to satisfy the high-resolution demands. At the other extreme, the less constrained {\em triclinic} crystallographic symmetry, requires the prediction of all three lengths $a$, $b$, $c$ and angles $\alpha$, $\beta$, and $\gamma$, and hence, needs exhaustive stepping through a six-dimensional space. To do this, domain knowledge is used to limit the range of each parameter, and the required resolution is represented as a pre-defined step size. Labelled samples are then generated by simulating the Bragg profiles at known values of the six parameters within these ranges by stepping through the parameter space in the pre-defined step sizes along all the size dimensions.
In practice, stepping through this high-resolution grid is very time-consuming, especially since the less constrained symmetries that require stepping through higher dimensional grids also take more time for each simulation, resulting in a data generation process that can span a few hours to days. 

Although increasing the amount of training data typically leads to better model performance, exhaustive traversal of structured grids can overwhelm the training process with uninformative examples. The primary motivation of this work is to address this challenge by designing an efficient workflow that incorporates an active learning (AL) approach. This approach aims to reduce the total number of samples needed for training, compared to traditional supervised learning, while maintaining model performance and providing quantified uncertainty.

AL integrates data generation and training in the same computational workflow and replaces the task of having to simulate at each grid point with an interactive query process. This query process selects a subset of samples to simulate, based on an adaptive criterion or policy that seeks to reduce the error and uncertainty in the predictions. 

Another motivation is to design an efficient streaming AL-based workflow with better resource utilization to train ML models using large volumes of labelled samples (high memory demands) on modest parallel heterogeneous (CPU+GPU) computing hosts which are generally more readily accessible for most practitioners.

%%%%%%%%%%%%%%%%%%%%%%%%%%%%%%%%%%%%
\subsection{Related Work}
\label{sec:related}

ML methods have been used in neutron scattering data for various tasks such as (1) using auto-encoders  to effectively extract spin Hamiltonians~\cite{natcomm2020}, (2) to predict neutron scattering cross-sections to constrain the parameters of a pre-existing model Hamiltonian using principal component analysis with an artificial neural network~\cite{aps-mar-2019-1}, and (3) to study phase transitions in single crystal x-ray diffraction data with unsupervised ML approach~\cite{aps-mar-2019-3}. Similarly, ML-based approaches are gaining steady acceptance as a classification tool for the study of local chemical environment of specific metal families~\cite{aps-mar-2019-2}
 and to understand neutron physics \cite{hey2020}. 
 %\textcolor{red}{Does this sentence look correct?}
 Recent efforts to determine the parameters of neutron scattering experimental data using ML~\cite{bigdata2019} and deep learning~\cite{Cardona-Big-Data-2020, wang2022using} have been shown to be effective in practice. High-performance computing and the scientific community have been
building workflows with both simulation and ML-aided data analysis modules~\cite{exalearn} for applications ranging from materials~\cite{pederson2022machine}, electrical grid simulations~\cite{dong2020smart} to protein folding~\cite{jia2020pushing}.  The streaming pipeline reported here requires allocation of core-level CPU resources to enable parallel execution of tasks within a node while also leveraging NUMA locality to optimize performance, a feature that existing middleware systems, such as RADICAL-Cybertools~\cite{merzky2015radical}, Colmena~\cite{ward2021colmena}, and SmartSim~\cite{PARTEE2022101707}, do not currently offer. Gaussian process regression-based methods have been reported as a solution to experimental design in autonomous and optimal data acquisition while conducting experiments~\cite{noack2021gaussian, mcdannald2022fly}. More recently, AL methods have been applied in neutron spectroscopy to study magnetic and lattice excitations~\cite{natcomm2023}. To our knowledge, the use of AL methods to train structure-finding ML models in neutron diffractometry has not been reported in the literature.
%

%%%%%%%%%%%%%%%%%%%%%%%%%%%%%%%%%%%%
\subsection{Contributions and Organization of the Paper}
\label{sec:contrib}
The main contribution of this paper is the first reported use of AL policies to train structure-finding models in neutron diffractometry with {\em significantly reduced training data without any loss of accuracy} while {\em simultaneously improving the training time}. More specifically, we report:
\begin{itemize}
    \item an in-depth performance study of a new stream-based training pipeline that orchestrates simulation-, training- and AL-tasks on CPU+GPU systems to efficiently train structure-finding ML models for neutron diffractometry.
    \item  the ability of the AL policy to train structure finding models with $\sim$75\% reduction in the amount of training data {\em without} any loss of accuracy. 
    \item  the design and performance of a new CPU+GPU-based streaming pipeline that also improves the training time by $\sim$20\% compared to conventional approaches. 
\end{itemize}

These contributions are accomplished in three overall steps. In a first step, we design an AL-policy for our application (\mysection{al}). To study its efficacy, we use an ML model already performance-tested for this application~\cite{Cardona-Big-Data-2020} and generate training datasets using the GSAS-II simulator (\mysection{model}). In a second step, we implement the AL policy into what we call a {\em serial} training workflow and confirm that this new workflow trains the model with significantly smaller training dataset sizes than a conventional ({\em baseline}) workflow {\em and} without any loss of accuracy. In a third step, we design a {\em streaming} workflow with superior CPU+GPU resource utilization and demonstrate that it outperforms the serial workflow in total execution time while delivering the same advantages of the serial workflow over the baseline workflow. These workflows are discussed in \mysection{wf}. Finally, we present detailed experimental results %on two CPU+GPU heterogeneous systems 
in \mysection{results} and conclude in \mysection{conclusions}. 

%%%%%%%%%%%%%%%%%%%%%%%%%%%%%%%%%%%%
\section{Active Learning Policy}
\label{sec:al}

We propose a batch-mode active learning (AL) policy based on uncertainty sampling, where training data is selected from a distribution that favors examples with high model uncertainty. See \cite{lewis1994w} for relevant terminology. Our method uses variance reduction techniques to estimate the prediction variance at each input, aiming to select examples that minimize the estimator's variance. The novelty lies in integrating heteroscedastic uncertainty estimation models of \cite{garcia2021uncertainty} for identifying areas of parameter space where the model produces high uncertainty predictions. For a fixed neural network that predicts unit cell parameters and an associated heteroscedastic uncertainty estimate, we construct a probability distribution over the input space that factors both model uncertainty and a user-defined prior. By iterating this process, we create an AL policy that can be applied to unit-cell parameter estimation workflows.

Suppose we are trying to approximate the estimator $\EXP(y|x)$ of an output $y \in \cY$ given an input $x \in \cX$ via an ML model $\hat{y}(x;\cD)$ trained on some data $\cD \subset \cX \times \cY$. In our case, we are solving the inverse problem $\EXP(y|x) = S^{-1}(x)$ where $x$ is a Bragg profile, $y$ is a vector of unit-cell parameters and $S: \cY \to \cX$ can be realized as simulator of Bragg profiles. The error in the approximation between the true value $y$ and the model is
\begin{align}\label{eqn:eq_error1}
    \sigma^2(x;\cD) &:= \EXP[\|y-\hat{y}(x;\cD)\|^2 \big| x,\cD] 
\end{align}
where the expectation is meant to be with respect to the conditional distribution of $y$ given $x$, and $\|\cdot\|$ denotes Euclidean norm on $\cY$. The error $\sigma^2(x;\cD)$ includes the aleatoric uncertainty in the observations and the ineffectiveness of the model to predict $y$ given $x$ due to epistemic uncertainty and incompleteness of $\cD$. Our approach estimates and reduces the \textit{total expected uncertainty of the prediction} defined as
\begin{equation}\label{eqn:def_sigma2x}
    \sigma^2(\cD) := \int_\cY  \sigma^2(S(y);\cD) p_{\cY}(y) \ud y
\end{equation}
where $p_{\cY}$ is a prior probability distribution over $\cY$. We achieve this by using $\sigma^2$ to define a probability distribution over $\mathcal{Y}$ from which a new training set of parameters is sampled. We then show that a model trained on this new set will have lower total uncertainty.

We use a model that predicts, along with $\hat{y}(x;\cD)$, a heteroscedastic uncertainty estimation $\hat{\sigma}^2$ for $\sigma^2$. See \cite{garcia2021uncertainty}. We require $y \mapsto \hat{\sigma}^2(S(y),\mathcal{D})$ to be strictly positive and bounded over the set $\cY$ which is assumed compact. Thus, in principle, we could define
\begin{equation}\label{eqn:eq_pcD1}
    p(y) \propto \hat{\sigma}^2(S(y);\cD) \, p_{\cY}(y)
\end{equation}
and obtain a distribution that assigns high probability to examples with high estimated uncertainty. The prior distribution $p_{\cY}$ is included in \eqn{eq_pcD1} in order to avoid selecting examples that have high uncertainty but are not representative of the natural population, removing the problem of learning from outliers \cite{kumar2020active}.

Drawing samples from $p$ can be done in various ways by rejection sampling \cite{hormann2004automatic}. Direct rejection sampling from $p$ in \eqn{eq_pcD1} will require a number of trials that is proportional to $\hat{\sigma}^{-2}(\cD)$ which cannot be known in advance. 
In order to avoid this, we turn to an interpolated version of $\hat{\sigma}(\cdot;\cD)$ over $\cY$. Consider a pre-determined `study set' $\cS =\{\bar{y}_n\}_{n=1}^{N_0} \subset \cY $ of approximately equally-spaced unit cell parameters on which the right hand side of \eqn{eq_pcD1} can be reconstructed by means of interpolation. The set $\cS$ could be a grid mesh over $\cY$, for example. We propose to use a Gaussian mixture distribution as factor, and define
\begin{equation}\label{eqn:eq_pcD}
    p(y) \propto p_{\cY}(y) \sum_{n=1}^{N_0}  \frac{\hat{\sigma}^2(S(\bar{y}_n);\cD)}{\sum_m \hat{\sigma}^2(S(\bar{y}_m);\cD)} e^{-\frac{1}{2} \left(\frac{y - \bar{y}_n}{\tau}\right)^2}
\end{equation}
for some $\tau > 0$ comparable to the spacing between examples in $\cS$. If the study set is adequately equally spaced and the spread factor $\tau$ is well-chosen, then $p$ defined by \eqn{eq_pcD} is a probability density function such that its corresponding pushforward $S_\# p$ assigns more probability to areas of $\cX$ where the estimated uncertainty is high. Equal spacing of $\cS$ is important because otherwise the sum in  \eqn{eq_pcD} will artificially accumulate more mass around points of $\cS$ that are closer together than around isolated points, even if the uncertainty is the same. 

The active learning policy samples data across both the continuous parameter space and the discrete class space allowing it to effectively address data imbalance (e.g., from less common symmetry classes) or noisy/low-quality data.
%%%%%%%%%%%%%%%%%%%%%%%%%%%%%%%%%%%%
\section{Data and ML Model}
\label{sec:model}

In this study, we consider three of the seven possible crystallographic symmetry classes, namely, cubic, trigonal and tetragonal classes. The simplest class is the cubic class defined by a single regression parameter $a=b=c$ and $\alpha=\beta=\gamma=90\degree$. The trigonal symmetry class is defined by $a=b=c$ and $\alpha = \beta = \gamma \neq 90\degree$ while the tetragonal symmetry class is defined by $a=b\neq c$ and $\alpha = \beta = \gamma = 90\degree$. Using barium titanate as the test material, the sampling ranges of these cell parameters, presented in Table.~\ref{tab:exp-setup-detail}, were guided by domain knowledge about the material from prior research.
%\textcolor{red}{TIANLE: Their detailed choice can be found in Table.~\ref{tab:exp-setup-detail}, Column E2.} 
Each diffraction pattern $X$ is a set of 2807 2-tuples $(x,~I(x))$ where $x$ is the time-of-flight (ToF) and $I(x)$ is the GSAS-generated scattering profile~\cite{toby-2013-gsas2} using ToF in the range [1,360$\mu$s, 18,919$\mu$s] with a step size of 0.0009381$\mu$s.

The ML model $\mathcal{M}$ used in the training task of the workflows studied in this paper is a multitask network first reported in \cite{Cardona-Big-Data-2020}. The choice of the ML model is based on its superior predictive performance as reported in \cite{Cardona-Big-Data-2020}. It uses the output from the first fully connected layer of the deep neural network classifier, also described in \cite{Cardona-Big-Data-2020}, to train both a regressor and a classifier. The classifier updates the weights using the error obtained from cross-entropy loss while the regressor uses the regression loss in~\eqn{exalearn_loss-2} for every batch. The regressor outputs the predicted lattice parameters, and modified to also output an estimate for the heteroscedastic uncertainty $\log \hat{\sigma}^2(x,\mathcal{D})$. In summary, we choose the loss function to be:
\begin{equation}
\label{eqn:exalearn_loss-1}
    \mathcal{L}(y,\sigma;\cD) = \mathcal{L}_\textrm{class} + \mathcal{L}_\textrm{reg}
\end{equation}
where $\mathcal{L}_\textrm{class}$ is the conventional cross entropy loss for classification~\cite{crossentropy2023} and $\mathcal{L}_\textrm{reg}$ is the following regression loss that includes the heteroscedastic uncertainty \cite{garcia2021uncertainty}:
\begin{equation}
\label{eqn:exalearn_loss-2}
    \mathcal{L}_\textrm{reg} = \frac{1}{N} \sum_{n=1}^N \frac{\|y_n - y(x_n))\|^2}{\sigma^2(x_n)} + \log(\sigma^2(x)))
\end{equation}

Workflow  presented in Sec.~\ref{sec:results} also use the following conventional definition of mean squared error (MSE) as another metric of performance comparison:
\begin{equation}
\label{eq-mse}
    \textrm{MSE} = \frac{1}{N} \sum_{n=1}^N \|y_n - y(x_n))\|^2
\end{equation}
%%%%%%%%%%%%%%%%%%%%%%%%%%%%%%%%%%%%
\section{Workflow Description}
\label{sec:wf}

In this section, we explain the three workflows used in this work, namely, the baseline workflow (\mysection{basewf}), %(Fig.~\ref{fig_basic_workflow})
the serial workflow (\mysection{serialwf}) %(Fig.~\ref{fig_serial_workflow}), 
and the streaming workflow (\mysection{streamwf}). %(Fig.~\ref{fig:stream_workflow_theory} and \ref{fig:stream_workflow_actual}). 
All three workflows include at least one simulation task ($S$) and one training task ($T$). The simulation task is an MPI task that executes GSAS-II on the CPU (because the GSAS-II simulator is a CPU-only code).  The simulation task computes the Bragg profiles in {\em batches} of input parameters sampled from the space spanned by the unit cell parameters within ranges as specified in \mysection{model}. The simulation task, therefore, converts a batch of input parameters (labels) into an equal number of scattering profiles (training data). The training task uses PyTorch DDP, which allows it to train the ML model presented in Sec.~\ref{sec:model} on GPUs with data parallelism. It trains for a fixed number of epochs using the training data set and evaluates the model performance using the validation data after each epoch, saves the optimal model during the training, and its final performance is evaluated using the test data.

\subsection{Baseline Workflow}\label{sec:basewf} 
The baseline workflow is the conventional workflow that consists of two tasks, namely: (a) a (bulk) simulation task, which uniformly samples parameters in the parameter space, and simulates three sets of bulk data, viz., the training set, the validation set, and the test set, and (b) a (bulk) training task, which trains/evaluates a model with all data generated from the simulation task. This conventional workflow which does not include an AL policy is used as the baseline for assessing the effectiveness of our AL policy (see Sec.~\ref{sec:exp-al-vs-bulk}).

\subsection{Serial Workflow}\label{sec:serialwf} 
%\paragraph{\bf Serial Workflow} 
The serial workflow executes in multiple phases. It begins with phase 0 in which a simulation task $S_0$ uniformly samples input parameters in the parameter space, and simulates four sets of bulk data, namely, the training set $D_{T0}$, the validation set $D_{V}$, the test set $D_{T}$, and the study set $D_{S}$. The data sets $D_{V}$, $D_{T}$ and $D_{S}$ are kept unmodified and serve as the validation set, test set, and study set for the entire duration of the serial workflow. The training task $T_0$ in this phase trains/evaluates the ML model with the training/validation/test set, and saves the optimal model $M0$. In phase 1, the active learning task from Sec. \ref{sec:al} is followed using model $M0$, a uniform distribution over $\cX$, and the study set $D_{S}$. Namely, a new batch of input parameters $P1$ is sampled from the distribution $p$ in \eqn{eq_pcD}. The simulation task $S_1$ of the next phase then simulates a new batch of training data, $D_{T1}$, using $P1$ as the input batch of parameters. The training task $T_1$ then trains/evaluates the ML model $M0$ with $D_{T0}\cup D_{T1}$, $D_{V}$, and $D_{T}$, and saves the optimal model $M1$ in phase 1. This process is repeated in the subsequent phases of the workflow.

\begin{figure}
    \centering
    \includegraphics[scale=0.28]{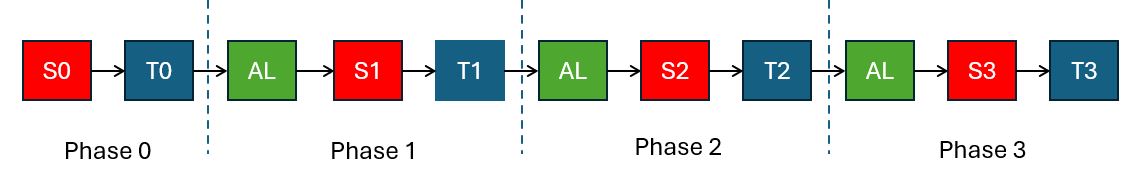}
    \vspace{1mm}
    \caption{Illustration of the serial workflow with four phases. 
    }
    \label{fig_serial_workflow}
    \vspace{-4mm}
\end{figure}

One drawback of the serial workflow is its low effective CPU and GPU utilization, since the simulation tasks only use CPU resources, and the training tasks mostly use GPU resources.
To enhance resource utilization in the workflow, one approach is to design a system where simulation and training tasks overlap, and the size of the training data is managed to ensure the execution times of the CPU-based simulations and GPU-based training tasks are well-balanced. However, this approach becomes non-trivial with the inclusion of an AL task due to the strict ordering required of the simulation, training and AL tasks for correct execution. In phase $k$, the training task $T_k$ relies on the training set $D_{Tk}$ simulated from the simulation task $S_k$, but to simulate data based on an AL policy, $S_k$ relies on the model obtained from the training task $T_{k-1}$ of the previous phase. Another potential pitfall in the computational performance of this workflow is that the model trains on larger and larger amounts of data as the number of phases in the workflow increases. However, the AL policy is expected to generate less redundant data with increasing phases potentially requiring less time to train the model at each successive phase of the workflow. As such, despite the increase in the size of the training data, the model is expected to require less number of epochs to train (for some fixed loss or accuracy) in this workflow. This serial workflow which includes the AL policy is used as the baseline for measuring the performance improvement of the new workflow introduced in the next subsection. 

\subsection{Streaming Workflow}\label{sec:streamwf} 
\begin{figure}
    \centering
    \includegraphics[scale=0.28]{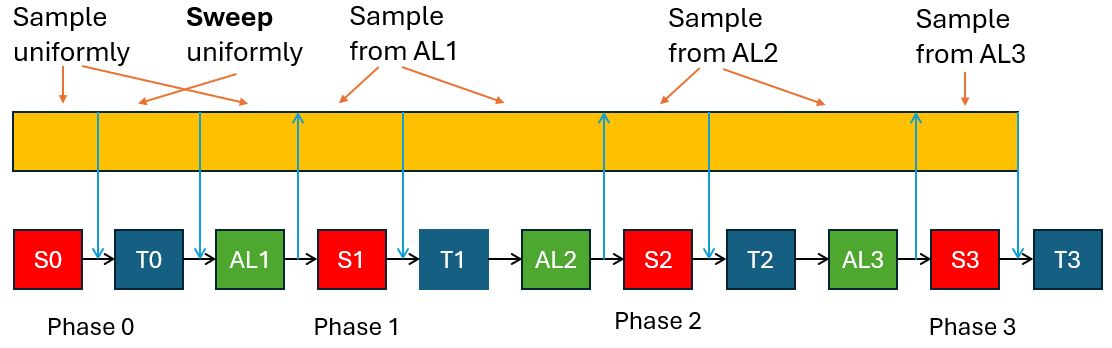}
    %\vspace{2mm}
    \caption{Ideal streaming workflow with four phases.
    }
    \label{fig:stream_workflow_theory}
    \vspace{-7mm}
\end{figure}

To overcome poor effective CPU and GPU utilization of the serial workflow, we present a pseudo-streaming workflow (which we still refer to as streaming workflow) to mimic an ideal streaming workflow. We will briefly describe the ideal streaming workflow for context and discuss the limitations that motivated the design of the pseudo-streaming workflow used in this study. 
In an ideal streaming workflow, two pipelines execute concurrently. Using a fixed amount of computational resources, the first pipeline, called the simulation pipeline, would continually generate data in a streaming fashion during the entire duration of the training campaign while the second pipeline, called the analysis pipeline, would execute like the serial pipeline but using fewer computational resources. \fig{stream_workflow_theory} shows this ideal streaming workflow. 

The simulation pipeline consists of a single simulation task that keeps executing on a {\em fixed subset of CPU resources} and generates a stream of Bragg profiles using cell lengths and cell angles as inputs sampled from the parameter space according to a probability distribution. The analysis pipeline is similar to the serial workflow in Fig.~\ref{fig_serial_workflow}. It uses the {\em remaining CPU resources} for portions of the simulation tasks and all available GPU resources for the training tasks. 
The two pipelines communicate when the simulation pipeline simulates a predefined fixed amount of training data which is transferred to the analysis pipeline for the model to train on. Similarly, the analysis pipeline communicates the most recently computed AL policy (a probability distribution) to the simulation pipeline for it to generate the next batch of training data.

In \fig{stream_workflow_theory}, the yellow-colored portions of the simulation pipeline that overlap with the red-colored simulation tasks of the analysis pipeline share {\em all available CPU resources}. The GPU-intensive training tasks in the analysis pipeline, shown by the blue-colored boxes, overlap with the CPU-only simulation tasks in the corresponding portions of the simulation pipeline. Similarly, the AL tasks in the analysis pipeline, shown by the green-colored boxes, are CPU-only and share the total number of available CPU resources with the corresponding overlapping portions of the simulation pipeline. This (ideal) streaming workflow maximizes the effective CPU usage during an AL-enabled training campaign.

\begin{figure}
    \centering
    \includegraphics[scale=0.28]{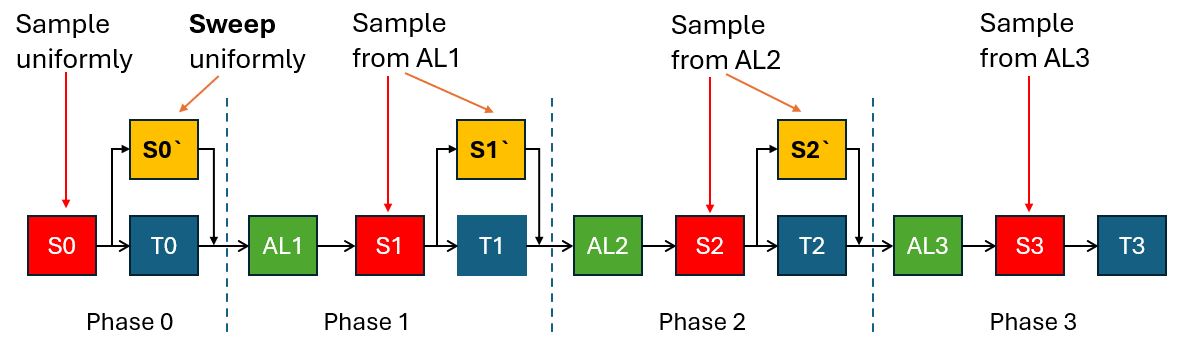}
    %\vspace{2mm}
    \caption{Streaming workflow with four phases.
    }
    \label{fig:stream_workflow_actual}
    \vspace{-4mm}
\end{figure}

In practice, the GSAS-II based simulation task does not generate streaming data but outputs the training samples in batches. To accommodate such a bulk production of simulated data, a pseudo-streaming workflow was designed to mimic the ideal streaming workflow. This pseudo-streaming workflow, which we will still refer to as the streaming workflow for ease of presentation, is shown in \fig{stream_workflow_actual}. The pseudo-streaming workflow is to be interpreted as follows: instead of a single simulation pipeline (yellow bar in Fig.~\ref{fig:stream_workflow_theory}), the simulation task can be thought of as divided into multiple smaller tasks. A small task that overlaps with a simulation task of the analysis pipeline can be thought of as merged into a single simulation task ($S_i$) that utilizes all the available CPUs.  On the other hand, a task ($S_i^\prime$) that overlaps with a training task ($T_i$) running on GPUs is executed concurrently on the available CPUs. The AL tasks ($AL_i$) use all the available CPUs and account for only a small (but constant) portion of the total training time. While not truly streaming, this pseudo-streaming workflow closely mimics the ideal streaming workflow and delivers significant improvements in its effective CPU usage over the serial workflow.

The streaming workflow begins with phase 0 in which a simulation task $S_0$ uniformly samples input parameters in the parameter space, and simulates three instead of four sets of bulk data, namely, the training set $D_{T0}$, the validation set $D_{V}$, and the test set $D_{T}$. The training task $T_0$ in this phase trains/evaluates the ML model with the training/validation/test set, and saves the optimal model $M0$. Concurrently, the simulation task $S_0^\prime$ simulates the study set $D_{S}$. This simulation is done using equally spaced input parameters (sweeps).
Hereafter, each new phase begins with an AL task. Informed by the AL policy computed by $AL_1$, a new batch of input parameters $P1$ is sampled from the distribution $p$ (see \eqn{eq_pcD}). The simulation task $S_1$ then simulates a new batch of training data, $D_{T1}$, using half of $P1$ as the input batch of parameters. The training task $T_1$ then trains/evaluates the ML model $M0$ with $D_{T0}\cup D_{T1}$, $D_{V}$, and $D_{T}$, and saves the optimal model $M1$ in phase 1. Concurrently with task $T1$, the simulation task $S_1^\prime$ simulates a new batch of training data, $D_{T1}^\prime$, using the other half of $P1$ as the input batch of parameters. This process is repeated in the subsequent phases of the workflow, with two differences, namely, (a) in phase $k$, the training task $T_k$ trains/evaluates the model not only with $D_{T_i}$ but also with $D_{T_i}^\prime$, and (b) in the last phase, the simulation task $S$ simulates the entire batch (not half) of input parameter $P$.

%%%%%%%%%%%%%%%%%%%%%%%%%%%%%%%%%%%%
\section{Performance Results}
\label{sec:results}

This section reports the performances of the baseline, serial, and streaming workflows. Four phases are considered for the serial and the streaming workflow with AL. 

\vspace{-2mm}
\subsection{Computing Testbeds}
\label{sec:testbed}
Performance of the three workflows was studied on two computing platforms, namely, the Polaris supercomputer in the Argonne Leadership Computing Facility (ALCF) and the Perlmutter supercomputer in the National Energy Research Scientific Computing (NERSC). Polaris is a 560 node HPE Apollo 6500 Gen 10+ based system. Each node has a single 2.8 GHz AMD EPYC Milan 7543P 32 core CPU with 512 GB of DDR4 RAM and four NVIDIA A100 GPUs connected via NVLink, a pair of local 1.6TB of SSDs in RAID0 for the users to use, and a pair of Slingshot network adapters. Perlmutter is a Cray EX supercomputer, a heterogeneous system with 1536 GPU accelerated nodes with 1 AMD Milan processor and 4 NVIDIA A100 GPUs, and 3072 CPU-only nodes with 2 AMD Milan processors interconnected using a 3-hop dragonfly network topology.

\vspace{-2mm}
\subsection{Experiment Settings}
\label{sec:expsets}
The size of the validation set $D_V$ and the testing set $D_T$ are kept equal to half the size of $D_{T0}$ (so that after the last phase, the relative size among training, validation, and testing is approximately $8:1:1$). The size of the study set $D_S$ is the same as that of $D_{T0}$. In the serial workflow, the number of samples in training set $D_{T0}$, $D_{T1}$, $D_{T2}$, and $D_{T3}$ are all the same. In the streaming workflow, the size of $D_{T1}$, $D_{T2}$, $D_{T1^\prime}$, $D_{T2^\prime}$ are 0.6 times that of $D_{T0}$, and the size of $D_{T3}$ is the same as that of $D_{T0}$. The number of epochs in each phase is approximately inversely proportional to $\sqrt{N_{tot}}$, where $N_{tot}$ is the total number of training samples in that phase. 
The hyperparameters are not fine-tuned but even without hyper-parameter tuning, the serial workflow with AL reduces the amount of data for training by a factor of four while improving the model accuracy. Additionally, the streaming workflow delivers about $13\%-24\%$ improvement over the serial workflow. Together, hyperparameter tuning has the potential to register even better performance.

\subsection{Performance of Simulation Tasks}
%\label{sec:exp-task-separate}
\label{sec:sim-task}

\begin{figure}[t]
%\vspace{-4mm}
\centering
\includegraphics[width=2.75in]{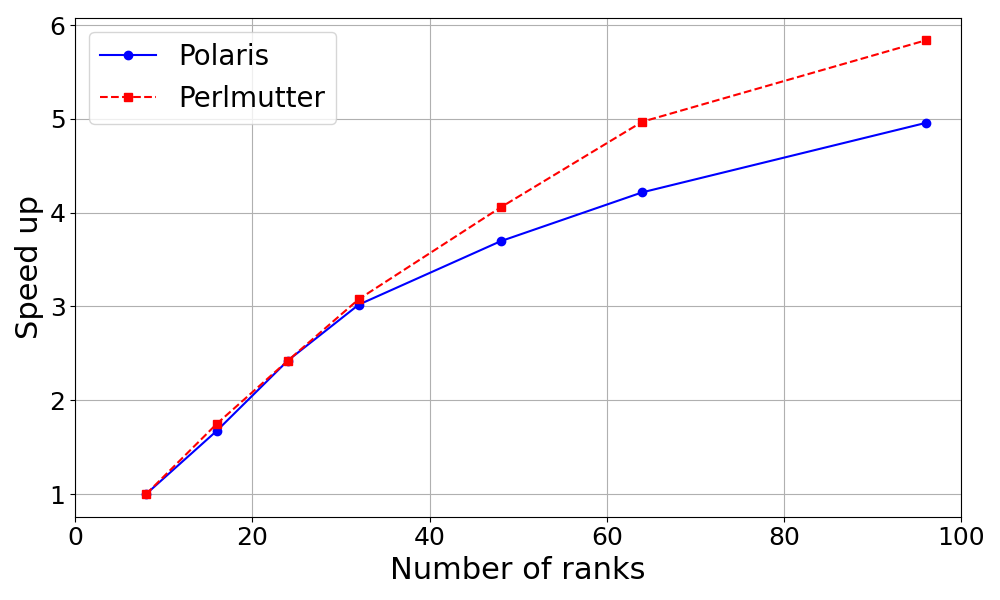}
\caption{The speed up (strong scaling) of simulation task.
} 
\label{fig:sim-scale}
\vspace{-4mm}
\end{figure}

Fig.~\ref{fig:sim-scale} shows the strong scaling behavior of the simulation task used to generate 13,500 samples. Due to prohibitively slow single-core execution, we opted for 8-core execution as the baseline. With fewer than 32 CPU cores, the speed-up is nearly linear
indicating effective strong scaling. However, as the number of CPU cores increases beyond this point, the speed-up no longer scales linearly. This deviation is attributed to the relatively small problem size, which allows the completion of the baseline case within a short wall time. For performance on instances with larger problem sizes, we direct the reader to \mysection{exp-serial-vs-stream-e2}.

\subsection{Performance of Training Tasks}
\label{sec:train-task}

\begin{figure}[ht!]
\vspace{-2mm}
\centering
\includegraphics[width=2.75in]{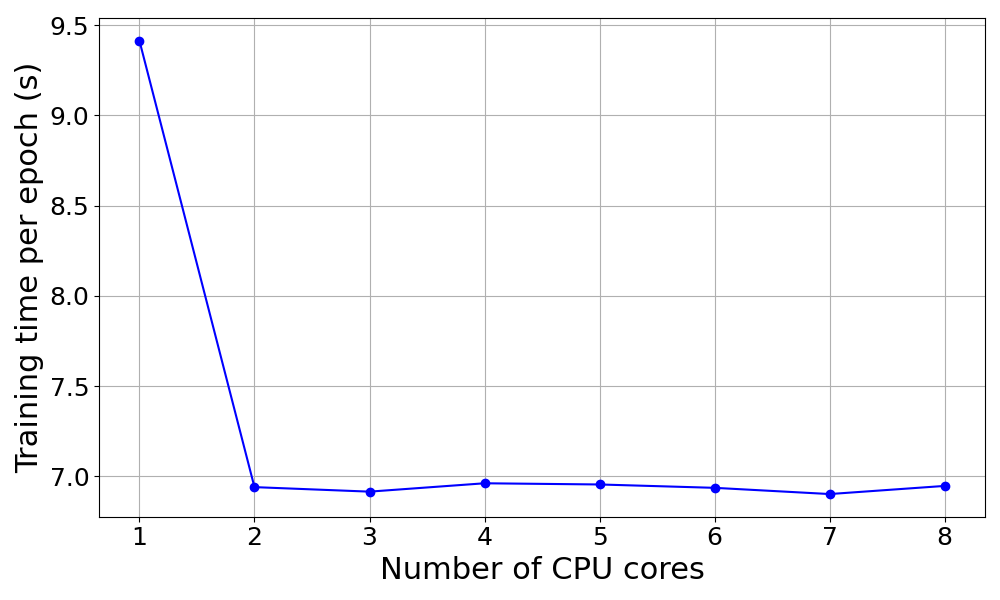}
\caption{The average training time per epoch when we vary the number of CPU cores used. Only a single process is used here.} 
\label{fig:train-cpu-thread}
\vspace{-4mm}
\end{figure}

Although the training tasks are performed on the GPU, each training process requires at least one CPU core to initiate and manage tasks executed on the CPU. In the streaming workflow, where simulation and training tasks run concurrently, it is crucial to balance the allocation of CPU cores between these tasks to minimize the total execution time. We maintain a fixed number of ranks and GPUs (one each) for training and adjust the number of CPU cores dedicated to this task to explore how different CPU core counts influence the execution time. The results, depicted in Fig.~\ref{fig:train-cpu-thread}, show a significant reduction in time (about $36\%$) when the number of CPU cores per process is increased from one to two. Further increases in CPU cores lead to a performance plateau. Consequently, in the streaming workflow, we allocate two CPU cores (with a single GPU) per training rank, while the remaining CPU cores (24 on Polaris, 56 on Perlmutter) are allocated for simulation tasks.

We should also be mindful of the NUMA (Non-Uniform Memory Access) domain when binding CPU cores for training tasks. Non-optimized CPU binding can lead to scenarios where, for example, four training ranks utilizing two CPU cores each might reside within a single NUMA domain, despite their associated GPUs being in different NUMA domains. We present the execution time performance of the training task under three CPU-binding configurations in Table~\ref{tab:train-numa}, namely: (a) NUMA-unfriendly, where all eight CPU cores are within the same NUMA domain, (b) NUMA-friendly-mismatch, where the eight CPU cores are distributed across four NUMA domains but do not align with the GPUs' NUMA domains (e.g., rank 0 uses CPUs in NUMA domain 0 and a GPU in NUMA domain 3), and (c) NUMA-friendly-match, where each rank's CPUs and GPU are located within the same NUMA domain. The two NUMA-friendly setups demonstrate approximately a $27\%$ performance improvement over the NUMA-unfriendly setup. However, the performance difference between the matched and mismatched setups is negligible, within about $3\%$. Therefore, in the streaming workflow, we consistently bind CPU pairs $\{8k, 8k+1\}$ and GPU $(3-k)$ on Polaris, and bind CPU pairs $\{16k, 16k+1\}$ and GPU $(3-k)$ on Perlmutter for each local rank $k$ (zero-indexed)\footnote{Polaris binds CPU $[8k, 8k+7]$ with GPU $(3-k)$ on the same NUMA domain, and Perlmutter binds CPU $[16k, 16k+15]$ with GPU $(3-k)$ on the same NUMA domain
%see \url{https://docs.alcf.anl.gov/polaris/hardware-overview/machine-overview/}
}.

\vspace{-2mm}
\begin{table}[ht]
\caption{The average training time per epoch for three different NUMA setups. The training task has four ranks, with each rank using a single GPU and two CPU cores.}
\vspace{-1mm}
\centering
\begin{tabular}{| l | c | c |}
\hline
{\bf NUMA Setup} &  {\bf Time (Polaris)} &  {\bf Time (Perlmutter)}\\
\hline\hline
NUMA-unfriendly & 2.52  & 2.20 \\
\hline
NUMA-friendly-mismatch & 1.98 & 1.84  \\
\hline
NUMA-friendly-match &  1.94 & 1.86 \\
\hline
\end{tabular}
\label{tab:train-numa}
\vspace{-2mm}
\end{table}

\begin{figure}[ht!]
\vspace{-2mm}
\centering
\includegraphics[width=2.75in]{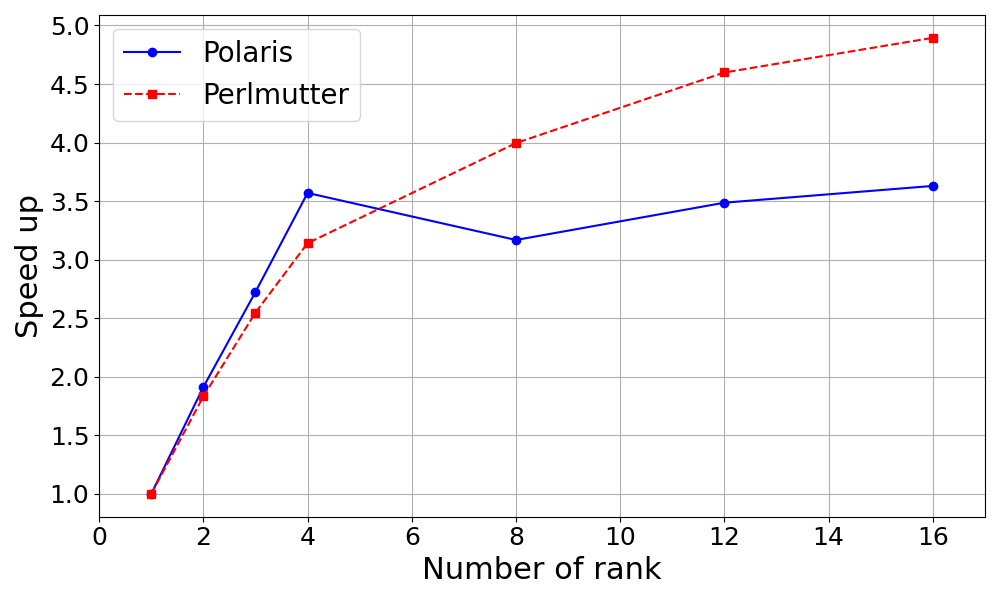}
\caption{Strong scaling speedup of training task for a fixed number of epochs.
%as we increase the number of ranks for the training. The total number of epochs is fixed, and we always use two CPU cores per rank. 
The CPU and GPU binding setup is NUMA-friendly-match.} 
\label{fig:train-gpu-scale}
\vspace{-4mm}
\end{figure}

Fig.~\ref{fig:train-gpu-scale} shows the strong scaling behavior of the training task. We observe near-linear scaling
as the GPU count increases from one to four. With further increase, the participating GPUs span multiple nodes resulting in sub-linear improvements as inter-node communication costs are larger than intra-node communication costs. For example, a four-process training configuration across four different nodes registers a performance decline of $45\%$ compared with a setup where all four ranks reside on the same node. These implications are further explored in Sec.~\ref{sec:exp-serial-vs-stream-e2}.

%\vspace{-4mm}

%\vspace{-4mm}
\subsection{Training Performance with AL}
\label{sec:exp-al-vs-bulk}

\begin{figure}[t]
%\begin{figure}[ht!]
%\vspace{-4mm}
\centering
\includegraphics[width=2.75in]{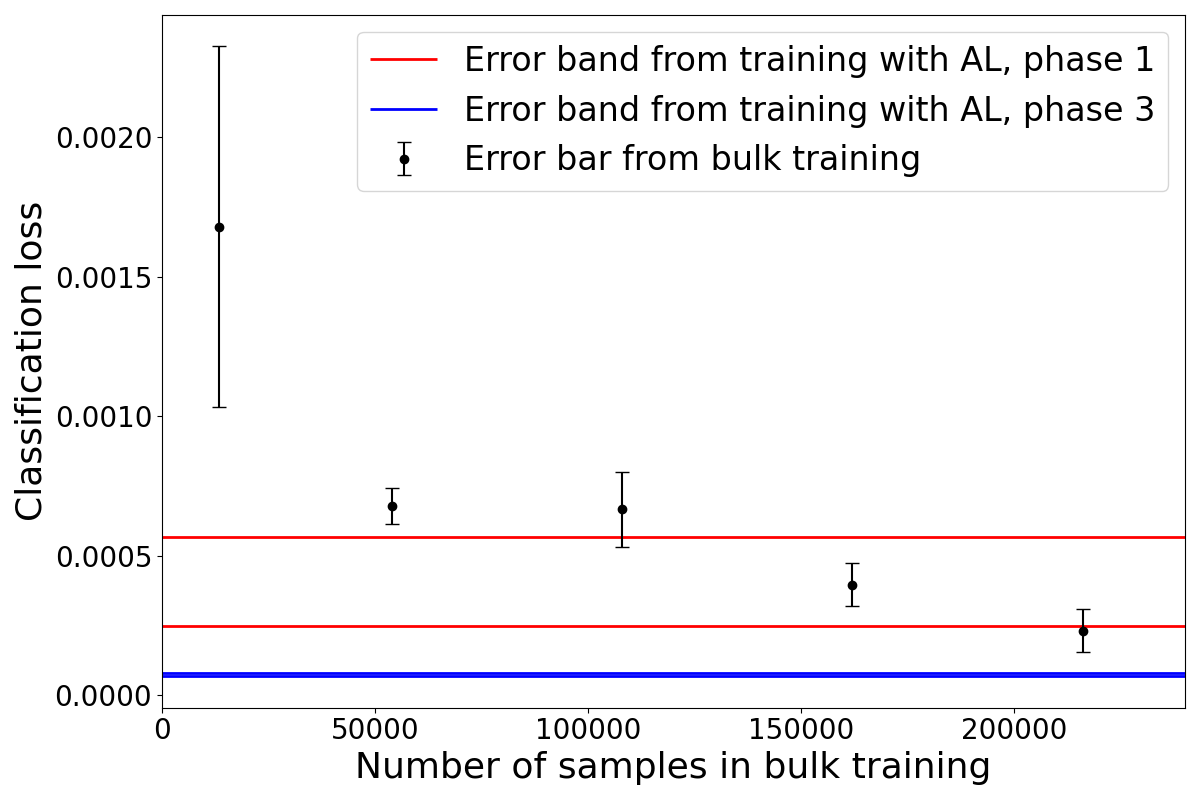}
\caption{Classification losses from AL workflow after phase 1 (red error band, which is trained with a total of 27000 samples), AL workflow after phase 3 (blue error band, which is trained with a total of 54000 samples), and the baseline workflow with different number of samples (see black error bar at different number of samples).} 
\label{fig:classification-loss-al-vs-bulk}
\vspace{-0.2in}
\end{figure}

In this section, we examine the impact of integrating the AL policy from \mysection{al} into our workflow by comparing the accuracy between models obtained with the baseline workflow and a four-phase AL serial workflow, as described in Sec.~\ref{sec:wf}. The comparison is made as follows. Each training task in the serial workflow trains the model on a dataset $D_{T_i}$ with 13,500 training samples in each phase $i$. The baseline workflow is executed with various sizes of the training dataset to determine the size of the dataset required to match the accuracy of the serial workflow. Fig.\ref{fig:classification-loss-al-vs-bulk} shows the classification loss on the test set from the serial workflow after phases one and three, together with those from the baseline workflow. 

\begin{figure}[h]
\centering
\includegraphics[width=2.75in]{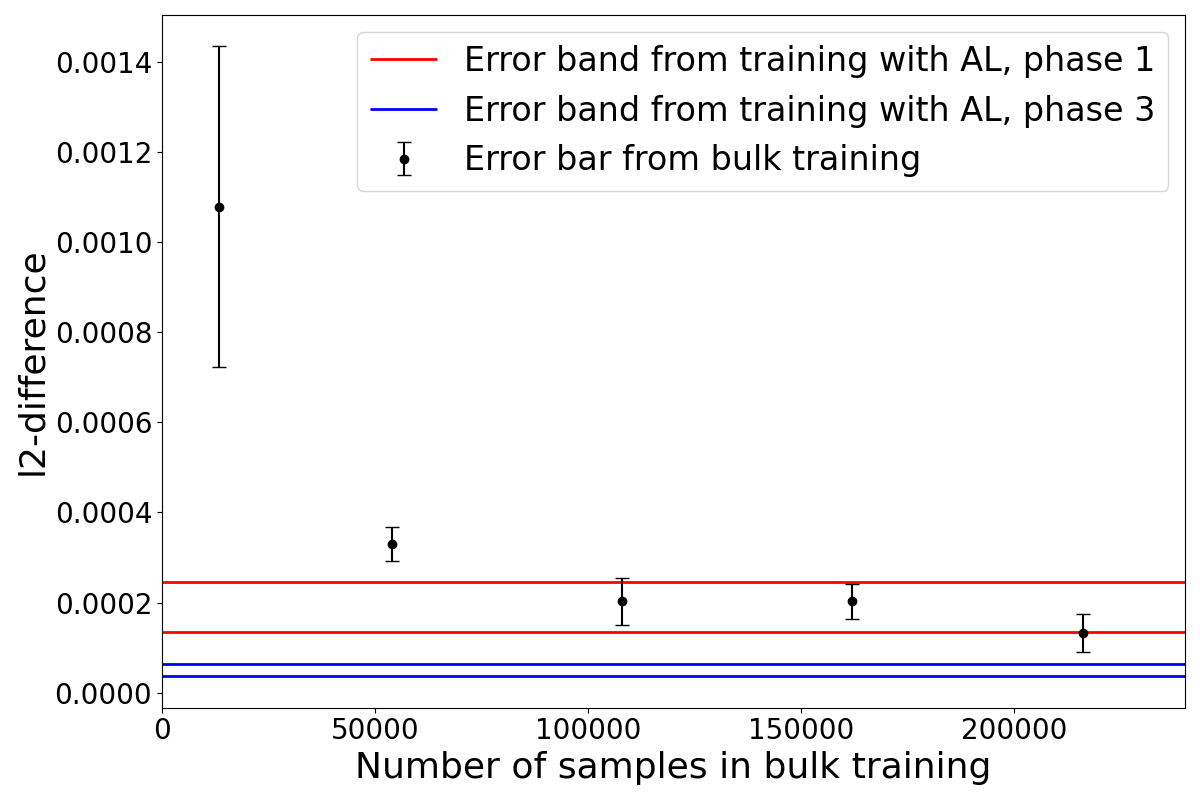}
\caption{Comparisons between the MSE from AL workflow after phase 1 (red error band, trained with a total of 27000 samples), AL workflow after phase 3 (blue error band, trained with a total of 54000 samples), and the baseline workflow with different number of samples (see black error bar at different number of samples).}
\label{fig:l2-difference-al-vs-bulk}
\vspace{-0.1in}
\end{figure}

To make the comparison robust, each workflow configuration was executed six times, each with a different seed, allowing us to plot error bands (the region between the top and bottom red (blue) lines defined by loss $\pm$ error) for statistical reliability. Note that in Fig.\ref{fig:classification-loss-al-vs-bulk}, the top and bottom blue lines are nearly overlapping. Error bars are also included for the baseline workflow to facilitate a clearer comparison with the AL-based serial training workflow. Classification losses from training with $n$ samples in the baseline workflow that are consistent (or worse/better) are those that are within the error band or (higher/lower) than the error band of the AL-based serial workflow. Similarly, Fig.\ref{fig:l2-difference-al-vs-bulk} shows the comparison of the MSE of the baseline workflow with the serial workflow with AL.

%\begin{figure}[ht!]

The results demonstrate that to achieve the same accuracy as the AL workflow in phase one (which is trained with 27,000 samples), the baseline workflow requires a training dataset that is four to six times larger. In addition, even when the training dataset of the baseline workflow is four times larger than that of the AL workflow, it significantly underperforms compared to the results of the AL workflow in phase three. This confirms that {\em the AL policy reduces the required number of training samples by more than a factor of four while achieving comparable levels of accuracy.} 

\vspace{-2mm}
\subsection{Serial Workflow vs. Streaming Workflow}
\label{sec:exp-serial-vs-stream}

In this section, we present the empirical results to evaluate and compare the accuracy and execution time performance of the serial and streaming workflows using two distinct datasets. Experiment E1 utilizes a relatively small dataset which allowed us to assess the effectiveness of the streaming workflow on a smaller scale. Experiment E2 involves a much larger dataset allowing us to vary the number of nodes used (one, two, and four), to yield more substantive results that confirm the  generalizability of our approach to larger datasets. In both experiments, training samples from only three of the seven possible crystallographic symmetry classes (see \mysection{intro}) were considered. The cell parameter ranges for these symmetry classes were chosen based on recommendations from domain experts. The setup details for E1 and E2 are in Table.~\ref{tab:exp-setup-detail}.

\vspace{-1mm}
\begin{table}[ht]
\caption{Experiment setups for E1 and E2.}
\vspace{-1mm}
\centering
\begin{tabular}{| p{4.5cm} | c | c |}
\hline
{\bf Parameter} & {\bf E1} & {\bf E2} \\
\hline
$a$ for cubic & $[3.5, 4.5)$ & $[2.5, 5.5)$ \\
\hline
$a$ and $c$ for trigonal/tetragonal & $[3.8, 4.2)$ & $[3.5, 4.5)$ \\
\hline
$\alpha$ for trigonal & $[60\degree, 120\degree)$  & $[30\degree, 120\degree)$ \\
\hline
\# training samples for {\em each symmetry} in $S_0$  & 4500 & 72000 \\
\hline
\# validation samples in $S_0$ & 6750 & 108000 \\
\hline
\# test samples in $S_0$ & 6750 & 108000 \\
\hline
\# study samples in $S_0$ ($S_0^\prime$) & 13500 & 216000 \\
\hline
\# samples generated in $S_1$, $S_2$ and $S_3$ in serial workflow, and in $S_3$ in streaming workflow & 13500 & 216000 \\
\hline
\# samples generated in $S_1$, $S_1^\prime$, $S_2$, $S_2^\prime$ in streaming workflow & 8100 & 129600 \\
\hline
batch size & 512 & 1024 \\
\hline
\# epochs in $T1$ & 400 & 400 \\
\hline
\# epochs in $T2$ & 300 & 300 \\
\hline
\# epochs in $T3$ & 250 & 250 \\
\hline
\# epochs in $T4$ & 200 & 200 \\
\hline
\end{tabular}
\label{tab:exp-setup-detail}
\end{table}

\subsubsection{\bf Experiment E1}
\begin{figure}[ht!]
%\vspace{-2mm}
\centering
\includegraphics[width=2.75in]{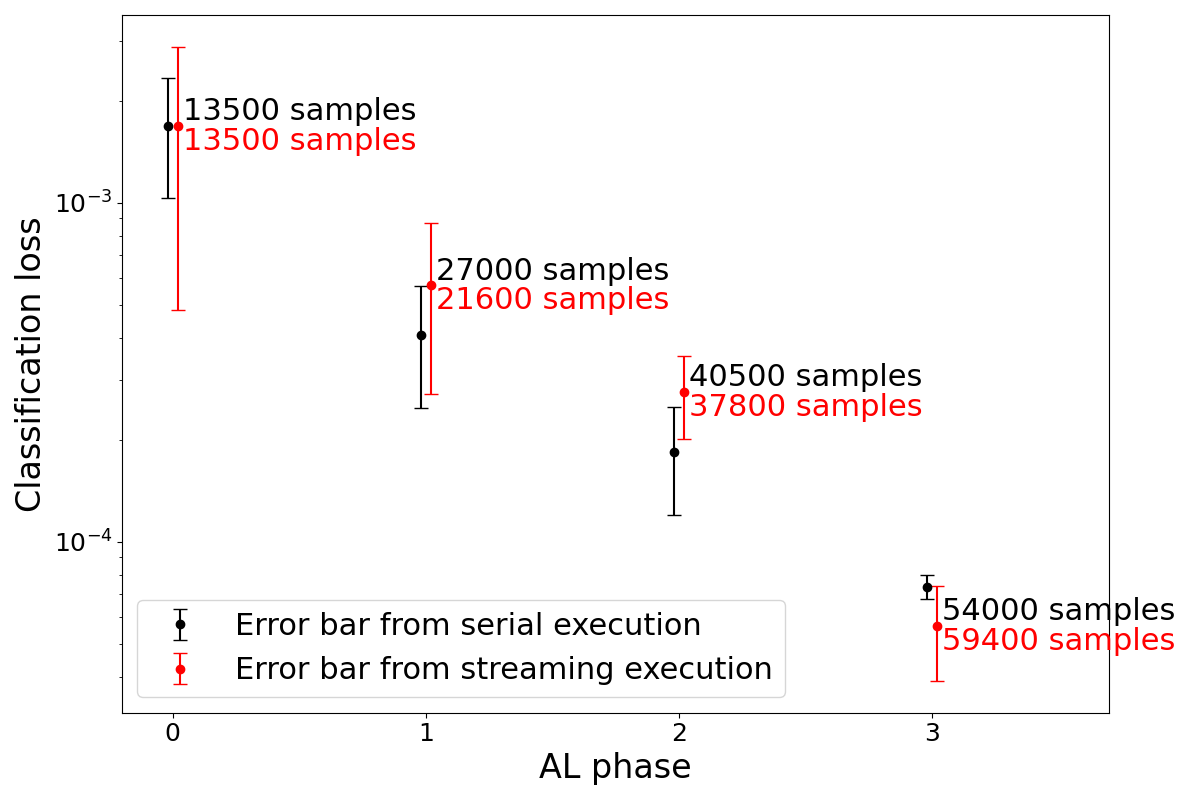}
\caption{Classification loss for serial and streaming workflows. The number of samples at each point is the size of the training dataset used in that phase.} 
\label{fig:classification-loss-serial-vs-stream}
\end{figure}

\begin{figure}[ht!]
\vspace{-4mm}
\centering
\includegraphics[width=2.75in]{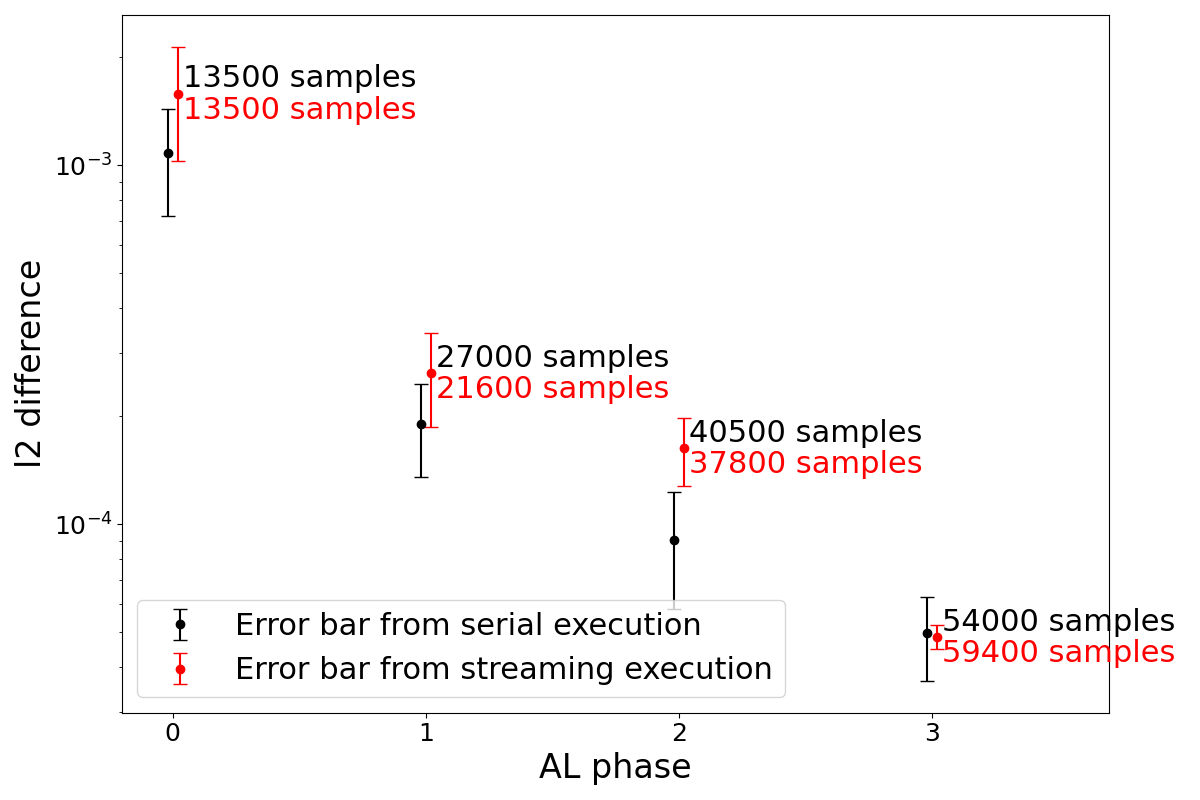}
\caption{MSE loss for serial and streaming workflows. The number of samples at each point is the size of the training dataset used in that phase.}
\label{fig:l2-difference-serial-vs-stream}
\vspace{-6mm}
\end{figure}

In this experiment, each workflow was executed six times on Polaris using different seeds to improve the robustness of the results. We compare the classification loss and MSE of the serial and streaming workflows across each phase, as illustrated in Fig.\ref{fig:classification-loss-serial-vs-stream} and Fig.\ref{fig:l2-difference-serial-vs-stream}. We also label the total number of training samples used in each phase for both workflows. The results indicate that although the serial workflow performs marginally better than the streaming workflow during the second and third phases, likely due to the higher number of samples used in those phases, the streaming workflow achieves comparable or even slightly superior performance in the final phase. This consistency confirms that the streaming workflow does not compromise accuracy performance relative to the serial workflow, demonstrating its effectiveness in maintaining accuracy while offering the benefit of reduced training time as shown next.

In Table~\ref{tab:exec-time-serial-vs-stream-e1}, we present the task-level execution times for the serial and streaming workflows. While the streaming workflow does not compromise accuracy performance, it improves the total execution time by approximately $19\%$. This improvement can be understood as follows. Firstly, task $S_0$ in the serial workflow requires 1.5 times longer to execute compared to its counterpart in the streaming workflow. In the serial workflow, $S_0$ includes the simulation of the training, testing, validation, and study datasets, while in the streaming workflow, the task of simulating the study set is shifted to $S_0^\prime$, which runs concurrently with the training task $T_0$. This task parallel execution reduces the overall execution time. Secondly, the intermediate simulation tasks, $S_1$ and $S_2$, are completed faster in the streaming workflow compared to the serial workflow. This is due to the fact that in the streaming workflow, the number of samples to be simulated in these two tasks is only 60\% of those in the serial workflow. Furthermore, the additional samples generated in $S_1'$ and $S_2'$ do not extend the overall duration of the workflow, as these tasks are executed in parallel with the training tasks $T1$ and $T2$, and these simulation tasks are completed faster than the concurrent training tasks. In summary, within Experiment $E1$, the streaming workflow outperforms the serial workflow in terms of speed, achieving a $19\%$ reduction in execution time using the same resources but with better utilization and without compromising the accuracy of the model. 

\begin{table}[t]
\caption{Execution time for each task in the serial and streaming workflows for $E1$. $\textrm{PG}_i$ stands for parallel group $i$ formed by $T_i$ and $S_i^\prime$ that run in parallel.}
%\vspace{2mm}
\centering
\begin{tabular}{| c | c | c |}
\hline
{\bf Task} & {\bf Serial (ms)} & {\bf Streaming (ms)} \\
\hline
S0 & $362887 \pm 23104$ & $234431 \pm 14970$ \\
\hline
T0 & $145536 \pm 14313$ & $167303 \pm 6889$ \\
S0$^\prime$ & -  & $162083 \pm 645$ \\
PG0 & -  & $167313 \pm 6889$ \\
\hline
AL & $4484 \pm 604$ & $4670 \pm 472$ \\
\hline
S1 & $99090 \pm 3459$ & $61244 \pm 6611$ \\
\hline
T1 & $119806 \pm 964$ & $123869 \pm 2479$ \\
S1$^\prime$ & -  & $86592 \pm 7904$ \\
PG1 & -  & $123883 \pm 2478$ \\
\hline
AL & $3756 \pm 148$ & $4531 \pm 170$ \\
\hline
S2 & $92114 \pm 6754$ & $55561 \pm 2844$ \\
\hline
T2 & $128138 \pm 1293$ & $124706 \pm 1508$ \\
S2$^\prime$ & -  & $76152 \pm 4308$ \\
PG2 &  - & $124721 \pm 1506$ \\
\hline
AL & $3630 \pm 45$ & $3821 \pm 80$ \\
\hline
S3 & $92625 \pm 3143$ & $82776 \pm 5481$ \\
\hline
T3 & $124286 \pm 972$ & $130927 \pm 1183$ \\
\hline
Total & $1184053 \pm 40316$ & $992944 \pm 26659$ \\
\hline
{\bf Speed up} & \multicolumn{2}{c|}{$\mathbf{1.19 \pm 0.05}$} \\
\hline
\end{tabular}
\label{tab:exec-time-serial-vs-stream-e1}
\vspace{-6mm}
\end{table}

\subsubsection{\bf Experiment E2}
\label{sec:exp-serial-vs-stream-e2}

\begin{table*}[ht]
\caption{Accuracy comparison of serial vs streaming workflows after each phase across different number of GPUs in $E2$.}
%\vspace{2mm}
\centering
\begin{tabular}{| c || c | c || c | c || c | c |}
\hline
{\bf \# GPUs} & \multicolumn{2}{c||}{{\bf 4}} & \multicolumn{2}{c||}{{\bf 8}} & \multicolumn{2}{c|}{{\bf 16}} \\
\hline
{\bf Metric} & {\bf Serial} & {\bf Streaming} & {\bf Serial} & {\bf Streaming} & {\bf Serial} & {\bf Streaming} \\
\hline\hline
Phase 0 & \multicolumn{6}{c|}{{-}} \\
\hline
MSE & $0.00124$ & $0.00149$ & $0.00125$ & $0.00125$ & $0.00133$ & $0.00145$ \\
\hline
Classification loss & $0.00642$ & $0.00495$ & $0.00634$ & $0.00424$ & $0.00778$ & $0.00516$ \\
\hline\hline
Phase 1 & \multicolumn{6}{c|}{{-}} \\
\hline
MSE & $0.000869$ & $0.000993$ & $0.000845$ & $0.000966$ & $0.000745$ & $0.000987$ \\
\hline
Classification loss & $0.00288$ & $0.00323$ & $0.00310$ & $0.00241$ & $0.00274$ & $0.00357$ \\
\hline\hline
Phase 2 & \multicolumn{6}{c|}{{-}} \\
\hline
MSE & $8.74\times 10^{-5}$ & $0.000122$ & $7.07\times 10^{-5}$ & $0.000105$ & $8.10\times 10^{-5}$ & $9.53\times 10^{-5}$ \\
\hline
Classification loss & $0.000205$ & $0.000288$ & $0.000265$ & $0.000382$ & $0.000148$ & $0.000294$ \\
\hline\hline
{\bf Phase 3} & \multicolumn{6}{c|}{{-}} \\
\hline
MSE & $1.12\times 10^{-5}$ & $1.38\times 10^{-5}$ & $1.02\times 10^{-5}$ & $1.02\times 10^{-5}$ & $1.10\times 10^{-5}$ & $9.73\times 10^{-6}$ \\
\hline
Classification loss & $2.99\times 10^{-5}$ & $2.83\times 10^{-5}$ & $2.58\times 10^{-5}$ & $3.58\times 10^{-5}$ & $3.40\times 10^{-5}$ & $3.55\times 10^{-5}$ \\
\hline
\end{tabular}
\label{tab:acc-serial-vs-stream-e2}
\vspace{-4mm}
\end{table*}

In this experiment, each workflow is executed only once since the data size is 16 times larger than in experiment $E1$ and requires significant computational resources to execute. Each workflow was tested on one, two, and four nodes (4 GPUs per node) of both the Polaris and Perlmutter supercomputer to evaluate their scalability. We compare the classification loss and MSE of both the serial and streaming workflows through each phase, as summarized in Table~\ref{tab:acc-serial-vs-stream-e2}. It indicates that the serial workflow generally outperforms the streaming workflow during phases 1 and 2 (except for the classification loss when the node count is 2), while in the final phase, the performance of the streaming workflow becomes comparable to that of the serial workflow. This is similar to the trend observed in $E1$. This consistent outcome confirms that the streaming workflow maintains accuracy performance comparable to the serial workflow, aligning with observations from $E1$.

\begin{table}[ht]
\caption{The total execution time of the serial and streaming workflow across different numbers of GPUs in $E2$, and the speed up between them.}
\centering
\begin{tabular}{|c|c|c|c|}
\hline
\textbf{Platform} & \textbf{Setup} & \textbf{Time (s)} & \textbf{Speed up} \\ \hline
\multirow{6}{*}{Polaris} & 4 GPUs - Serial & $18318.2$ & \multirow{2}{*}{\bf 1.24} \\ %\cline{2-3}
 & 4 GPUs - Stream & $14812.9$ &  \\ \cline{2-4}
 & 8 GPUs - Serial & $14062.0$ & \multirow{2}{*}{\bf 1.19} \\ %\cline{2-3}
 & 8 GPUs - Stream & $11840.4$ &  \\ \cline{2-4}
 & 16 GPUs - Serial & $10701.8$ & \multirow{2}{*}{\bf 1.13} \\ %\cline{2-3}
 & 16 GPUs - Stream & $9455.1$ &  \\ \hline
\multirow{6}{*}{Perlmutter} & 4 GPUs - Serial & 16523.2 & \multirow{2}{*}{\bf 1.22} \\ %\cline{2-3}
 & 4 GPUs - Stream & 13587.2 &  \\ \cline{2-4}
 & 8 GPUs - Serial & 12803.7& \multirow{2}{*}{\bf 1.15} \\ %\cline{2-3}
 & 8 GPUs - Stream & 11104.6 &  \\ \cline{2-4}
 & 16 GPUs - Serial & 10123.0 & \multirow{2}{*}{\bf 1.12} \\ %\cline{2-3}
 & 16 GPUs - Stream & 9073.5 &  \\ \hline
\end{tabular}
\label{tab:exec-time-serial-vs-stream-e2}
\vspace{-5mm}
\end{table}

In Table~\ref{tab:exec-time-serial-vs-stream-e2}, we present the total execution time of the serial and streaming workflows and their relative speed up on one, two, and four nodes (4, 8 and 16 GPUs, respectively). The streaming workflow reduces the total execution time from approximately 12\% to 24\%. The reason for this speedup is similar to that for $E1$. However, the speedup is seen to decrease with increasing node counts. The most important reason is the different scalability behaviors of the training and simulation tasks that emerge when training with large data sets as is in experiment $E2$. While the simulation task scales nearly linearly from one to four nodes, the training task does not. This difference leads to a less optimal balance between the execution times of these tasks in the three parallel groups as the number of nodes increases. Effective task balancing within parallel groups is crucial for achieving high speedup and optimal resource utilization. In the extreme case where one task considerably outlasts the other, resources allocated to the shorter-duration task remain idle when the other long-lived task is running. With effective task balancing within parallel groups, the execution time of the streaming workflow is expected to improve.

%%%%%%%%%%%%%%%%%%%%%%%%%%%%%%%%%%%%
\vspace{-1mm}
\section{Conclusions}
\label{sec:conclusions}
This paper presents the design and performance study of an efficient streaming pipeline to train a tested structure-finding ML model for neutron diffractrometry on two heterogeneous CPU+GPU computing platforms. It was shown to train the model with $\sim$75\% less training data without any loss of accuracy while also reducing the training time by $\sim$20\% compared to conventional training workflows due to the integration of a new AL policy, introduced here for  the first time, and the efficient streaming design of our workflow that maximized the use of available CPU and GPU resources on the computing hosts. In this study, three of the seven crystallographic symmetry classes were used. We plan to assess the performance of our streaming pipeline on additional symmetry classes beyond the three tested here. 

The training pipeline presented here is directly applicable to X-ray diffractometry. In addition, a streaming pipeline is being integrated into ongoing work involving deep encoder-decoder networks used as surrogates for diffusion equations\cite{toledo2023analyzing}. 
In this context, our pipeline shows potential for reducing the total execution time of model training with a different AL policy. 
Our uncertainty-based AL method is also being applied to the early stages of a project on the Theia detector design plan~\cite{askins2020theia}, where ML techniques are being employed to speed up the detector design. As these projects develop, we anticipate further confirmation of the generalizability of our approach to a wider range of application areas, particularly those involving simulation steering and large-scale scientific workflows. The code base for this paper can be accessed at 
\href{https://github.com/GKNB/ALBAND/tree/main}{https://github.com/GKNB/ALBAND/tree/main}.
%here: 
%\href{https://github.com/GKNB/ALBAND/tree/main}{\bf {\texttt{ALBAND}}}. 

\vspace{-1mm}
\section*{Acknowledgment}
The authors would like to acknowledge the contributions of Aymen Al-Saadi and Andrew Park of the Rutgers University, USA, and Tianle Wang of the Brookhaven National Laboratory, USA, in porting the active learning-based streaming pipeline presented in the original paper published in IEEE BigData 2024 to ROSE (RADICAL Optimal and Smart-Surrogate Explorer), a framework for supporting concurrent and adaptive execution of simulation and surrogate training tasks on HPC resources, that can be accessed at \href{https://github.com/radical-cybertools/ROSE/tree/main/examples/neutron-scattering}{https://github.com/radical-cybertools/ROSE/tree/main/examples/neutron-scattering}. These additional contributions were supported by the National Science Foundation under Grant No. NSF 2212549 (Al-Saadi and Park) and by the Department of Energy under the award DOE ASCR DE-SC0021352 (Wang).

%\vspace{-1mm}
\bibliographystyle{IEEEtran}
%\vspace{-1mm}
\bibliography{main}

%-----------------------------------------------------------------------------
\end{document}